\documentclass[10pt,journal,compsoc]{IEEEtran}
\usepackage{amsmath,amsfonts}
\usepackage{algorithmic}
\usepackage{algorithm}
\usepackage{array}
\usepackage[caption=false,font=normalsize,labelfont=sf,textfont=sf]{subfig}
\usepackage{xcolor}
\usepackage{textcomp}
\usepackage{stfloats}
\usepackage{url}
\usepackage{verbatim}
\usepackage{graphicx}
\usepackage{cite}
\usepackage{makecell}
\usepackage{multirow}
\usepackage{pifont}
\usepackage[colorlinks = true,
            linkcolor = blue,
            urlcolor  = blue,
            citecolor = blue,
            anchorcolor = blue]{hyperref}

\newcommand{\cmark}{\textcolor{green!60!black}{\ding{51}}} 
\newcommand{\xmark}{\textcolor{red}{\ding{55}}}

\begin{document}

\title{Rethinking Graph Super-resolution: Dual Frameworks for Topological Fidelity}

\author{Pragya Singh and Islem Rekik 
\thanks{P. Singh and I. Rekik are affiliated with BASIRA Lab, Imperial-X and Department of Computing, Imperial College London, London, UK. Corresponding author: Islem Rekik, Email: i.rekik@imperial.ac.uk, \url{https://basira-lab.com} and \url{https://ix.imperial.ac.uk/}}}



\maketitle

\begin{abstract}

Graph super-resolution, the task of inferring high-resolution (HR) graphs from low-resolution (LR) counterparts, is an underexplored yet crucial research direction that circumvents the need for costly data acquisition. This makes it especially desirable for resource-constrained fields such as the medical domain. While recent GNN-based approaches show promise, they suffer from two key limitations: (1) matrix-based node super-resolution that disregards graph structure and lacks permutation invariance; and (2) reliance on node representations to infer edge weights, which limits scalability and expressivity. In this work, we propose two GNN-agnostic frameworks to address these issues. First, Bi-SR introduces a bipartite graph connecting LR and HR nodes to enable structure-aware node super-resolution that preserves topology and permutation invariance. Second, DEFEND learns edge representations by mapping HR edges to nodes of a dual graph, allowing edge inference via standard node-based GNNs. We evaluate both frameworks on a real-world brain connectome dataset, where they achieve state-of-the-art performance across seven topological measures. To support generalization, we introduce twelve new simulated datasets that capture diverse topologies and LR-HR relationships. These enable comprehensive benchmarking of graph super-resolution methods. Our source code is available at \url{https://github.com/basiralab/DEFEND}.
\end{abstract}


\section{Introduction}
\IEEEPARstart{H}{igh-resolution} (HR) datasets are crucial for accurate analysis and information processing, but their acquisition is resource-intensive. This necessitates the development of super-resolution techniques to enhance the quality of easily accessible low-resolution (LR) datasets. Consequently, super-resolution has been extensively studied for images \cite{dong2015image,greenspan2009super,lu2019satellite,wang2022comprehensive}. However, many real-world datasets are inherently represented as graphs, such as molecular structures, brain connectivity, and social interactions. Yet, graph super-resolution remains underexplored. 

\begin{figure*}[!t]
\centering
\includegraphics[width=\textwidth]{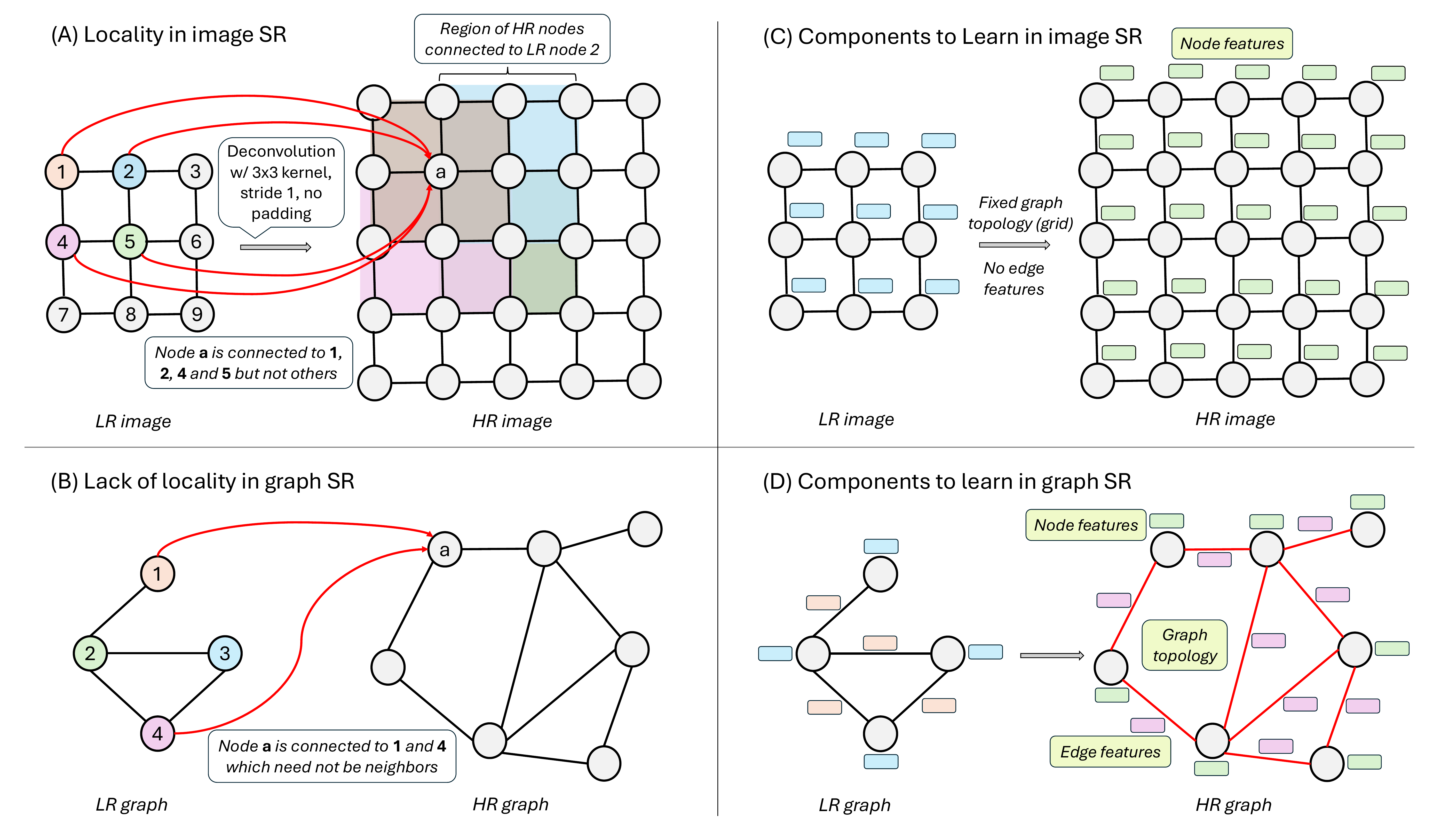}
\caption{Comparison of image and graph super-resolution (SR). (A) and (C) illustrate image SR as a special case of graph SR, where the graph topology is fixed to a regular grid and node features correspond to pixel values. (B) and (D) represent the general case of graph SR considered in this work, which involves learning both the topology and edge features in addition to node features, without assuming spatial locality between LR and HR nodes.}
\label{fig1}
\end{figure*}

Notably, image super-resolution can be seen as a special case of graph super-resolution where the graph topology is fixed (regular grid of predetermined size), and only node features (pixel intensities) need to be inferred. This grid structure also induces spatial locality, whereby each HR pixel is estimated from a local receptive field over the LR input. In contrast, graph super-resolution requires predicting both the graph topology and node/edge features, often without the LR-HR locality seen in image super-resolution. These differences make graph super-resolution significantly more challenging (see Figure \ref{fig1}). 

Nevertheless, graph super-resolution is highly relevant in domains like network neuroscience, where the connectivity between brain regions is represented as a brain graph, or connectome. HR connectomes have been shown to improve neural fingerprinting and behavior prediction \cite{tian2021high,hayasaka2010comparison,zalesky2010whole,finn2015functional,cengiz2019predicting}, however, their acquisition and processing are computationally intensive, and even small graphs may require gigabytes per individual \cite{tian2021high}. Graph super-resolution offers a scalable alternative by inferring them from LR connectomes.

\begin{figure}[!t]
\centering
\includegraphics[width=\columnwidth]{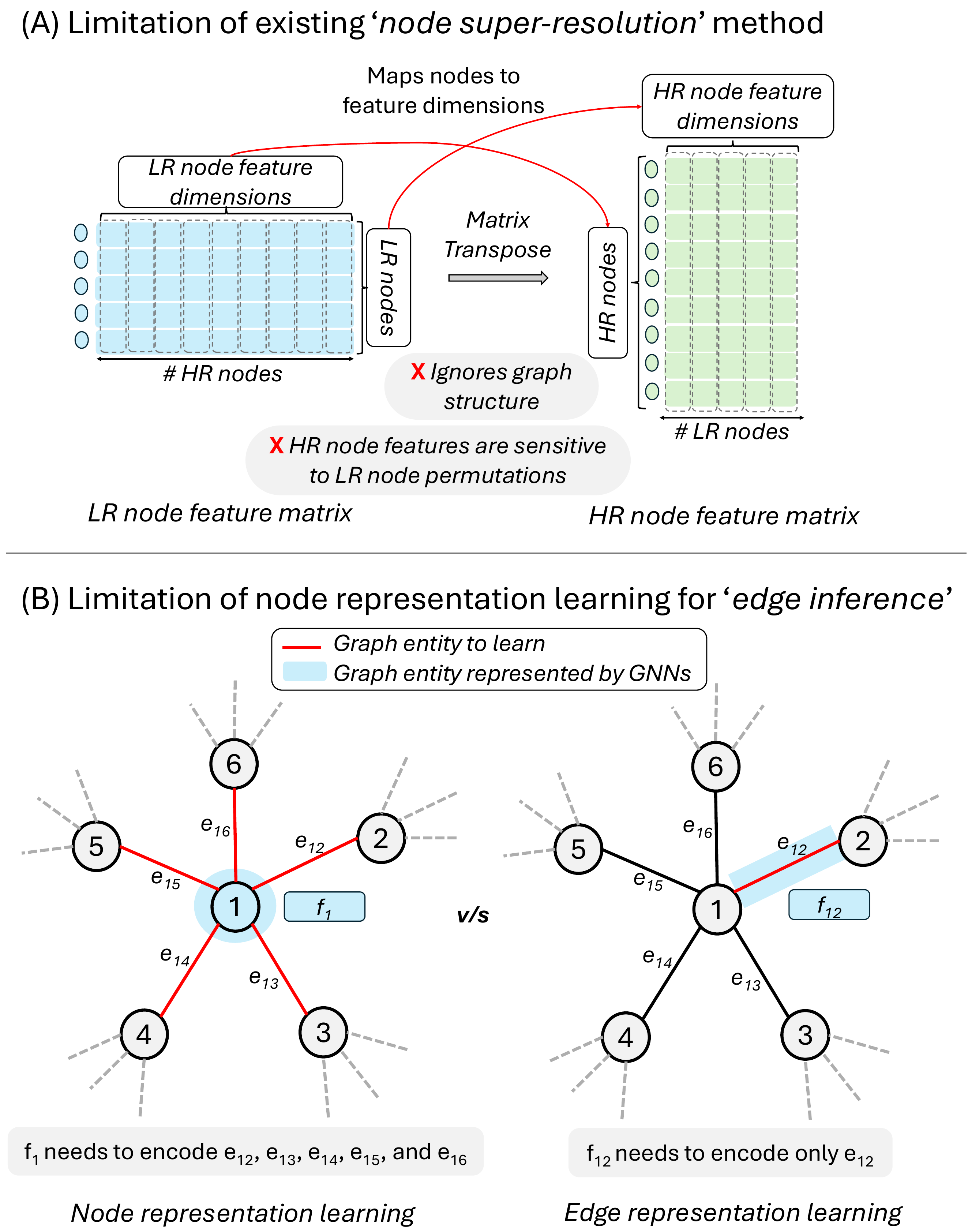}
\caption{Limitations of existing brain graph super-resolution methods. (A) Current approaches to \emph{node super-resolution} rely on a matrix transpose operation, which ignores graph structure and yields HR node features that are sensitive to LR node permutations. (B) GNN-based \emph{edge learning} uses node representations to encode all incident edges, making them computationally expensive and limiting its capacity to model edges.}
\label{fig2}
\end{figure}

Building on this application and the broader success of graph neural networks (GNNs) \cite{zhou2020graph,bronstein2017geometric,wang2021scgnn}, recent approaches adapt GNNs to brain graph super-resolution \cite{isallari2021brain,pala2021template,mhiri2021non}, typically comprising two stages: \emph{node super-resolution}, which infers HR node features from LR counterparts, and \emph{edge inference}, which predicts HR topology and edge features from node representations. Despite promising results, these approaches face two key limitations: (1) To infer HR node features, they rely on a simple linear algebraic technique (matrix transpose) that ignores graph structure and is sensitive to LR node permutations; (2) Since most GNNs rely on node representation learning, they use computationally expensive message-passing to learn node features capable of encoding all incident edges, offering limited scalability and capacity to model graph topology. Together, these limitations underscore critical structural gaps in current graph super-resolution methods (see Figure \ref{fig2}). 

The second limitation is especially consequential in network neuroscience, where the brain graph topology has been shown to play a central role in detecting neurodegenerative diseases such as Alzheimer’s (AD) and Parkinson’s (PD) \cite{pereira2015aberrant,pereira2016disrupted,khazaee2015identifying,nigro2022role,mijalkov2017braph}. For example, \cite{pereira2016disrupted} reports that AD progression correlates with reduced path length and mean clustering compared to a control group. \cite{pereira2015aberrant} observes aberrant values for clustering coefficient, characteristic path length, and small-worldness in 3T MRI data for early-stage PD patients. \cite{nigro2022role} finds that frontotemporal dementia involves both hub loss and compensatory hub emergence in distinct brain regions.

Motivated by the limitations of existing methods, we propose two GNN-agnostic frameworks that address graph super-resolution in a principled manner. (1) \textbf{Bi}partite \textbf{S}uper-\textbf{R}esolution (\textbf{Bi-SR}) introduces a bipartite graph connecting LR and HR nodes, enabling each HR node to aggregate information from all LR nodes. Unlike matrix transpose, this formulation preserves graph structure and is invariant to LR node permutations. (2) \textbf{D}ual graphs for \textbf{E}dge \textbf{FE}ature lear\textbf{N}ing and \textbf{D}etection (\textbf{DEFEND}) performs edge representation learning by leveraging a dual graph transformation that maps edges of the HR graph to nodes in a new dual graph. Node computations on this dual graph correspond naturally to edge computations on the HR graph, allowing for edge inference via simpler GNN architectures. Our models achieve state-of-the-art performance on brain graph super-resolution across seven key topological measures, demonstrating their utility for downstream neuroscientific analysis. To support broader applicability, we further benchmark these frameworks on a comprehensive suite of simulated datasets that span diverse graph topologies and LR-HR relationships.

\section{Related Work}

Although graph super-resolution remains an underexplored area, several foundational studies have contributed notable advancements. The method in \cite{isallari2021brain} proposed a graph U-Net architecture \cite{gao2019graph}, incorporating a hierarchical structure and graph Laplacian-based upsampling for LR brain graphs \cite{tanaka2018spectral}. The approach in \cite{pala2021template} accelerated training by using template graphs at both low and high resolutions as structural priors. In \cite{mhiri2021non}, NNConv layers \cite{simonovsky2017dynamic} were employed for global graph alignment, along with a graph-GAN model \cite{wang2018graphgan} to generate HR connectomes. However, this method often results in out-of-memory (OOM) errors on dense graphs due to the computational demands of NNConv layers. Lastly, \cite{monti2018dual} introduced a dual graph formulation for computing attention weights in GAT layers \cite{velickovic2017graph}, which differs from our use of dual graphs for direct edge inference in graph super-resolution.

\section{Preliminaries}

\textbf{Graph Representation.} Let $\mathcal{G} = (\mathcal{V}, \mathcal{E}, \mathbf{A}, \mathbf{X})$ denote a graph with node set $\mathcal{V}$ and edge set $\mathcal{E}$, where $n=|\mathcal{V}|$. The adjacency matrix is $\mathbf{A} \in \mathbb{R}^{ n\times n}$, with $\mathbf{A}_{ij} \in [0, 1]$. Let $\mathbf{X} \in \mathbb{R}^{n \times d}$ denote the node feature matrix, where each row $\mathbf{X} = \mathbf{x}_i \in \mathbb{R}^d$ is the feature vector of node $i$. 

\textbf{Message Passing Neural Networks (MPNNs).} This work considers a subclass of MPNNs \cite{gilmer2017neural} wherein node features are updated at each layer $l$ via:
\begin{align}
\mathbf{z}_i^{(l)} &= \beta^{(l)} \mathbf{x}_i^{(l-1)} + (1 - \beta^{(l)}) \sum_{j \in \mathcal{N}_i} w_{ij}^{(l)} \mathbf{x}_j^{(l-1)} \\
\mathbf{x}_i^{(l)} &= f_n(\mathbf{z}_i^{(l)})
\end{align}
where $\mathbf{x}_j^{(l-1)} \in \mathbb{R}^d$ denotes the feature vector of node $j$ at layer $l-1$, $\mathcal{N}_i$  is the set of neighbors of node $i$, $w_{ij}^{(l)} \in \mathbb{R}$ is a (fixed or learnable) weight that may depend on node features, and $\beta^{(l)} \in [0,1]$ controls the trade-off between self and neighborhood information. The function $f_n$ is learnable and assumed to be a universal approximator for theoretical analysis. This formulation subsumes several established architectures, including GCN~\cite{kipf2016semi}, GIN \cite{xu2018powerful}, GAT \cite{velickovic2017graph}, and Graph Transformers \cite{shi2020masked}.

For an $L$-layer architecture, we denote the output as $\mathbf{X}^{(L)} = GNN(\mathbf{X}^{(0)}, \mathbf{A})$. This mapping is permutation-equivariant \emph{s.t.} for any permutation matrix $\mathbf{P} \in \{0, 1\}^{n \times n}$, $GNN(\mathbf{P}\mathbf{X}^{(0)}, \mathbf{P}\mathbf{A}\mathbf{P}^T) = \mathbf{P}GNN(\mathbf{X}^{(0)}, \mathbf{A})$.

\textbf{Problem Statement.} Let $\mathcal{G}_l = (\mathcal{V}_l, \mathcal{E}_l, \mathbf{A}_l, \mathbf{X}_l)$ and $\mathcal{G}_h = (\mathcal{V}_h, \mathcal{E}_h, \mathbf{A}_h, \mathbf{X}_h)$ denote the LR and HR graphs, respectively, related by the transformation $\mathcal{G}_l = T_{\delta}(\mathcal{G}_h)$, where $T_{\delta}$ captures task-specific abstraction. The goal is to learn a super-resolution model $S_{\theta}$ that approximates $\mathcal{\hat{G}}_h = (\mathcal{V}_h, \mathcal{E}_h, \mathbf{\hat{A}}_h, \mathbf{\hat{X}}_h) = S_{\theta}(\mathcal{G}_l)$ by minimizing the reconstruction loss $\mathcal{L}(\mathcal{\hat{G}}_h, \mathcal{G}_h)$.

For the GNN-based $S_{\theta}$ considered in this work, we define $S_{\theta} = S_{\mathcal{E}} \circ S_{\mathcal{V}}$, where $\mathbf{\hat{X}}_h = S_{\mathcal{V}}(\mathbf{X}_l, \mathbf{A}_l)$ performs \emph{node super-resolution} and $\mathbf{\hat{A}}_h = S_{\mathcal{E}}(\mathbf{\hat{X}}_h)$ performs \emph{edge inference}. Existing work relies on matrix transpose $S_{\mathcal{V}}(\mathbf{X}_l, \mathbf{A}_l) = GNN_l(\mathbf{X}_l, \mathbf{A}_l)^T$, where $GNN_l: \mathbb{R}^{n_l \times d} \times \{0, 1\}^{n_l \times n_l} \mapsto \mathbb{R}^{n_l \times n_h}$, and dot-product $S_{\mathcal{E}}(\mathbf{\hat{X}}_h) = \mathbf{\hat{X}}_h \mathbf{\hat{X}}_h^T$. While the former ignores graph structure, the latter has limited capacity to model edges. We address these issues by formulating $S_{\mathcal{V}}$ using bipartite message passing to preserve graph structure and LR node permutation invariance, and by designing $S_{\mathcal{E}}$ via a dual graph transformation that enables expressive and efficient edge inference through an invertible mapping between HR edges and dual nodes.

\begin{table}[!t]
\caption{Comparison of graph super-resolution methods
\label{tab0}}
\centering
\renewcommand{\arraystretch}{1.2} 
\resizebox{\columnwidth}{!}{%
\begin{tabular}{|l|cccc|}
\hline
\textbf{Method} & \makecell[c]{Structure-aware\\node inference} & \makecell[c]{Permutation-invariant\\node inference} & \makecell[c]{Dedicated \&\\expressive edge inference} & \makecell[c]{GNN-\\agnostic} \\
\hline
SOTA \cite{mhiri2021non}  & \xmark & \xmark & \xmark & \xmark \\
Bi-SR & \cmark & \cmark & — & \cmark \\
DEFEND & — & — & \cmark & \cmark \\
Bi-SR + DEFEND & \cmark & \cmark & \cmark & \cmark \\
\hline
\end{tabular}%
}
\end{table}

\section{Proposed Bi-SR}\label{section:bi-sr}

\begin{figure*}[!t]
\centering
\includegraphics[width=\textwidth]{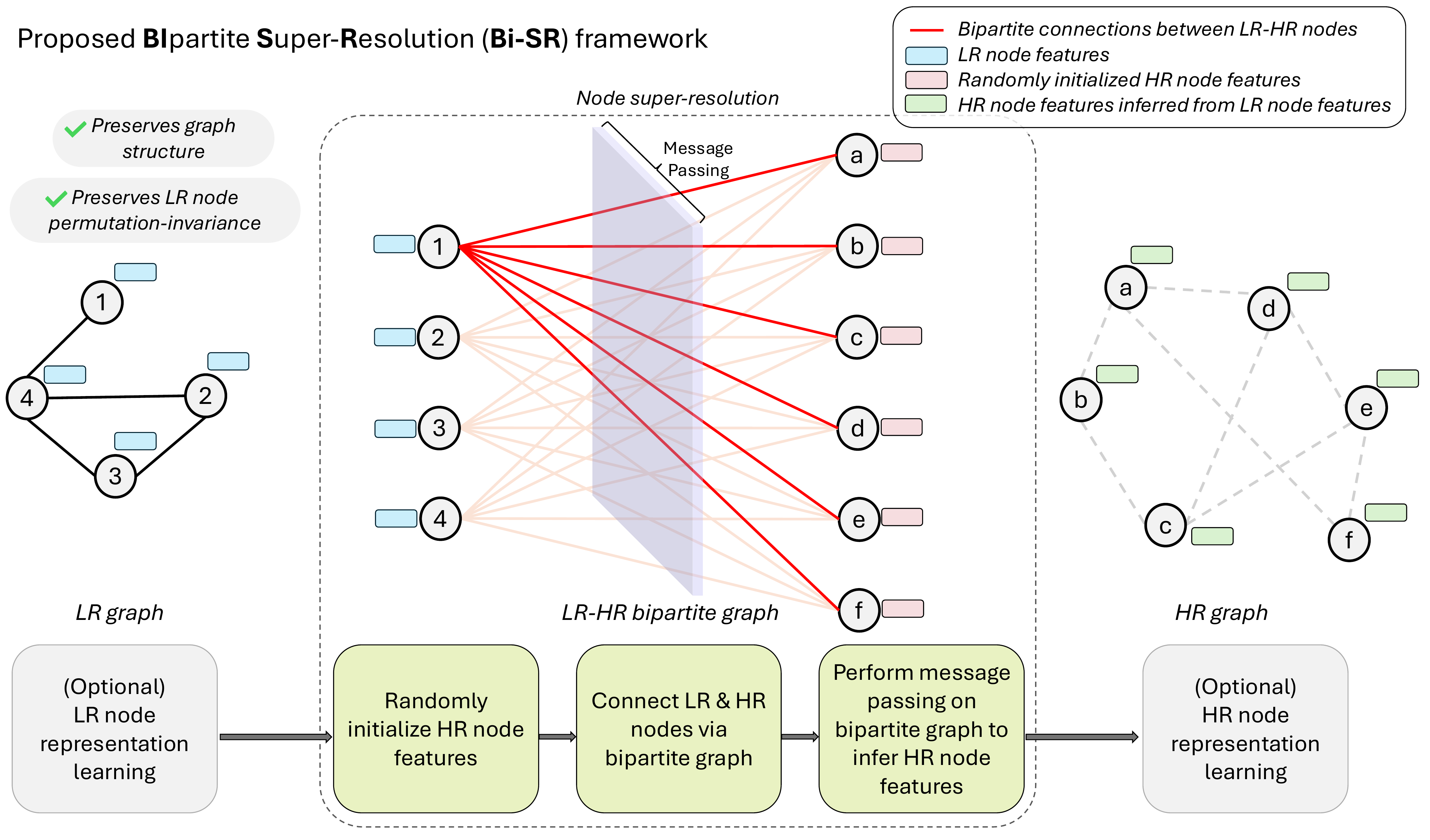}
\caption{Overview of the Bi-SR framework.
Given a low-resolution (LR) graph, Bi-SR constructs a bipartite graph between LR and high-resolution (HR) nodes, enabling each HR node to aggregate information from all LR nodes through message passing. HR nodes are initialized with fixed random features to break symmetry. Optional representation learning steps refine LR and HR features before and after bipartite propagation. This structure-aware formulation preserves the graph topology and ensures permutation invariance with respect to LR node ordering. }
\label{fig3}
\end{figure*}

We propose \textbf{Bi}partite \textbf{S}uper-\textbf{R}esolution (\textbf{Bi-SR}), a structurally-aware $S_{\mathcal{V}}$ framework for inferring high-resolution (HR) node features from a low-resolution (LR) graph. The method begins with optional refinement of LR features, followed by the initialization of HR node embeddings with random, fixed vectors to break symmetry. A fully connected bipartite graph is then constructed between LR and HR nodes, enabling message passing to propagate structural information and infer HR features. Optionally, these features are further refined through message passing over a computation domain defined among HR nodes, which may be fixed or learned. An overview of the complete framework is shown in Figure \ref{fig3}.

\textbf{LR node representation learning.} LR node features are optionally updated into a representation space more suitable for HR node inference as $\mathbf{X}_l = GNN_l(\mathbf{X}_l, \mathbf{A}_l)$. 

\textbf{HR node initialization.} HR nodes in the bipartite graph share the same neighborhood, inducing symmetry that hinders learning meaningful representations. To break this symmetry, we initialize the HR node features $\mathbf{X}_h^0 \in \mathbb{R}^{n_h \times d}$ by sampling i.i.d. entries from $\mathcal{U}(0, 1)$. By the law of large numbers \cite{hsu1947complete} and concentration of measure \cite{ledoux2001concentration}, these features have approximately constant norms and pairwise equidistance. $\mathbf{X}_h^0$ is kept fixed across all samples during training, ensuring unique and consistent encoding of HR nodes.

\textbf{Bipartite graph formulation.} Define the bipartite graph as $\mathcal{G}_b = (\mathcal{V}_b, \mathcal{E}_b, \mathbf{X}_b, \mathbf{A}_b)$, where $\mathcal{V}_b = \mathcal{V}_l \cup \mathcal{V}_h$, $\mathcal{E}_b = \{\{w, v\}| w \in \mathcal{V}_l, v \in \mathcal{V}_h\}$, $\mathbf{A}_b = \begin{bmatrix}
    \mathbf{0}_{n_l \times n_l} & \mathbf{1}_{n_l \times n_h}\\
     \mathbf{1}_{n_h \times n_l}& \mathbf{0}_{n_h \times n_h}
    \end{bmatrix}$, and $\mathbf{X}_b = \begin{bmatrix}
        \mathbf{X}_l \\ \mathbf{X}_h^0
    \end{bmatrix}$.

\textbf{HR node feature inference.} To infer $\mathbf{\hat{X}}_h$, we perform message passing on the bipartite graph as: 
\begin{equation}
    \mathbf{\hat{X}}_b = \begin{bmatrix}
\mathbf{X}_l^{'} \\ \mathbf{\hat{X}}_h
\end{bmatrix} = GNN_b(\mathbf{X}_b, \mathbf{A}_b)
\end{equation}
where, $GNN_b: \mathbb{R}^{(n_l+n_h) \times d} \mapsto \mathbb{R}^{(n_l+n_h) \times d'}$ and $\mathbf{X}_l^{'}$ is discarded. 

Alternatively, $\mathbf{\hat{X}}_h$ can be computed via linear combination on the bipartite graph: 
\begin{equation}
    \mathbf{\hat{X}}_h = \mathbf{W}^b\mathbf{X}_l
\end{equation}
where, $\mathbf{W}^b_{ij}$ denotes the learnable contribution of LR node $j$ to HR node $i$.

\textbf{HR node representation learning.} Optionally, $\mathbf{\hat{X}}_h$ can be refined by allowing HR nodes to interact via message passing. This requires defining a computation domain $\mathbf{A}_h^{ref} \in \mathbb{R}^{n_h \times n_h}$, distinct from the original HR adjacency matrix $\mathbf{A}_h$. For a fixed computation domain, we use full-connectivity: $\mathbf{A}_h^{ref} = \mathbf{1} - \mathbf{I}$. For a learnable computation domain (inspired by \cite{zaripova2023graph}), we first learn additional HR node features $\mathbf{X}_h^{ref}$ and compute:
\begin{equation}
    \begin{split}
        \mathbf{\hat{A}}_h^{ref} &= \sigma(\mathbf{X}_h^{ref} \cdot {\mathbf{X}_h^{ref}}^T)\\
        \mathbf{A}_h^{ref} &= \mathbf{\hat{A}}_h^{ref} \odot H(\mathbf{\hat{A}}_h^{ref} - 0.5)
    \end{split}
\end{equation}
where, $\sigma$ is the sigmoid function and $H(x)$ is the Heaviside step function(1 if $x \geq 0$, 0 otherwise). The final HR node features are then updated as $\mathbf{\hat{X}}_h = GNN_{ref}(\mathbf{\hat{X}}_h, \mathbf{A}_h^{ref})$. 

\textbf{LR node permutation invariance.} 
Let $\mathbf{P}_l$ be a permutation matrix acting on LR nodes. 
\begin{itemize}
    \item Matrix Transpose (\textbf{\emph{MT}}) method is not permutation invariant: 
    \begin{equation}
        GNN_l(\mathbf{P}_l\mathbf{X}_l, \mathbf{P}_l\mathbf{A}_l\mathbf{P}_l^T)^T = \mathbf{\hat{X}}_h^T\mathbf{P}_l^T \neq \mathbf{\hat{X}}_h
    \end{equation}
    \item Bipartite message passing (\textbf{\emph{Bi-MP}}) is permutation-invariant: The permutation matrix for the bipartite graph is obtained by permuting the LR nodes according to $\mathbf{P}_l$ while keeping the HR nodes unchanged, $\mathbf{P}_b = \begin{bmatrix}
            \mathbf{P}_l & \mathbf{0}\\
            \mathbf{0} & \mathbf{I}
        \end{bmatrix}$. Thus, 
        \begin{equation}
            GNN_b(\mathbf{P}_b\mathbf{X}_b, \mathbf{P}_b\mathbf{A}_b\mathbf{P}_b^T)=\mathbf{P}_b\mathbf{\hat{X}}_b =  \begin{bmatrix}
            \mathbf{P}_l\mathbf{X}_l^{'}\\
            \mathbf{\hat{X}}_h
        \end{bmatrix}
        \end{equation}
        implying that $\mathbf{\hat{X}}_h$ remains invariant.
    \item Bipartite linear combination (\textbf{\emph{Bi-LC}}) is not permutation invariant: 
    \begin{equation}
        \mathbf{W}^b(\mathbf{P}_l\mathbf{X}_l) \neq \mathbf{W}^b\mathbf{X}_l = \mathbf{\hat{X}}_h
    \end{equation}
\end{itemize}
As GNNs are node-permutation equivariant, composing above methods with additional GNNs for LR or HR node representation learning does not alter their permutation (non-)invariance.

While Bi-SR enables structure-aware and permutation-invariant inference of HR node features, it does not address edge inference. Recovering HR topology from these features requires a more expressive model than existing dot-product $S_{\mathcal{E}}$, motivating the design of our next framework DEFEND.

\section{Proposed DEFEND}

\begin{figure*}[!t]
\centering
\includegraphics[width=\textwidth]{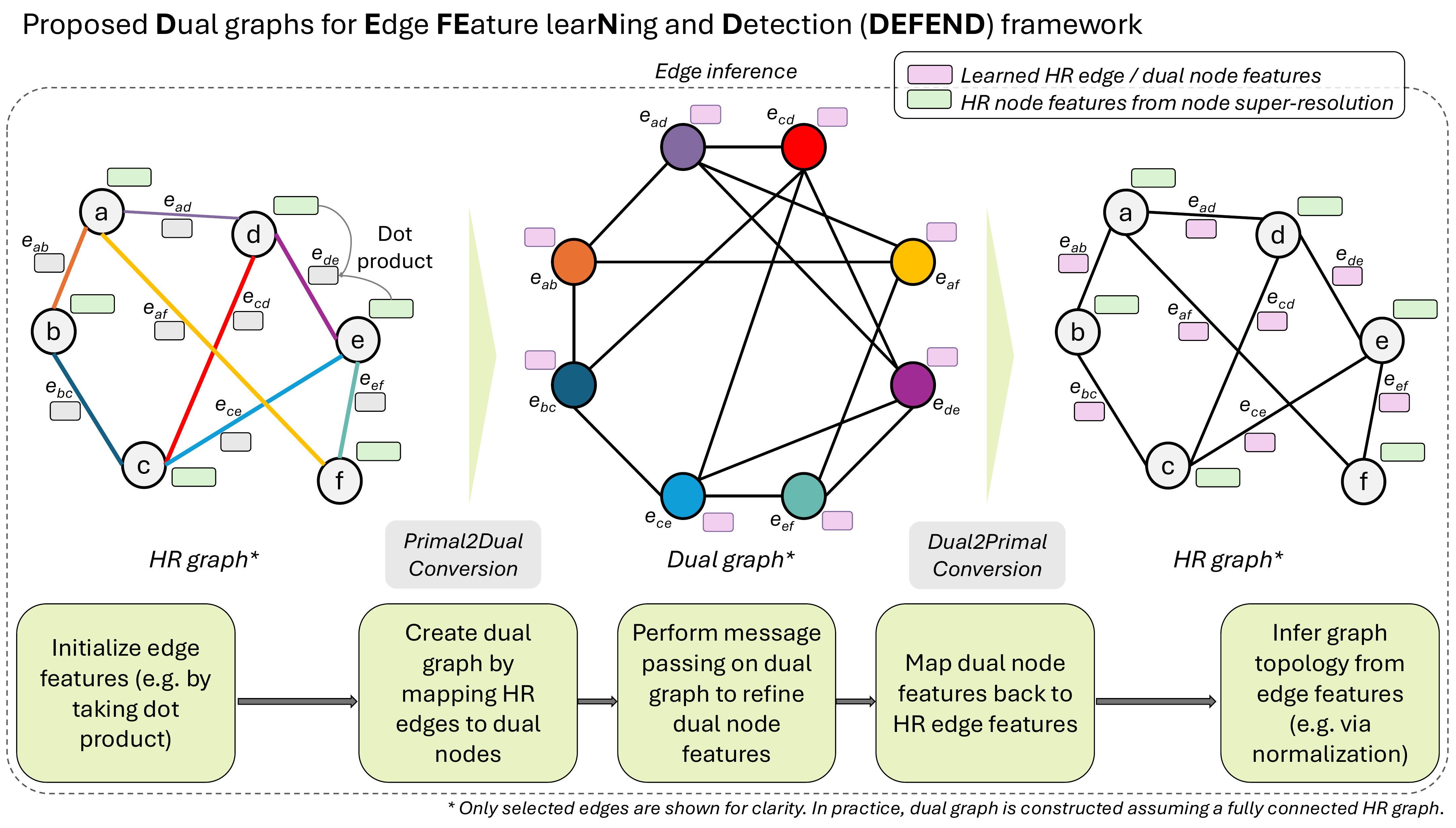}
\caption{Overview of the DEFEND framework.
Starting from HR node features, DEFEND initializes edge features (e.g., via dot product) and constructs a dual graph by mapping each HR edge to a dual node. Message passing is then performed on the dual graph to learn refined edge representations. These are mapped back to HR edges and used to infer the HR graph topology (e.g., via normalization). The primal-to-dual mapping is invertible and enables efficient edge inference using node-based GNNs.}
\label{fig4}
\end{figure*}

We propose \textbf{D}ual graphs for \textbf{E}dge \textbf{FE}ature lear\textbf{N}ing and \textbf{D}etection (\textbf{DEFEND}), a more expressive framework for learning edge features and inferring graph topology through a dual graph formulation. By creating bijective mapping between edges of our HR graph and nodes of a dual graph, DEFEND enables the use of standard node representation learning GNNs to learn edge representations. We begin by introducing the dual graph formulation, then describe each component of the framework, and conclude with its computational analysis. A schematic overview is shown in Figure \ref{fig4}.

\textbf{Dual Graph Formulation.} Given a simple undirected (primal) graph $\mathcal{G}_p = (\mathcal{V}_p, \mathcal{E}_p, \mathbf{A}_p)$, its dual graph is defined as $\mathcal{G}_d = (\mathcal{V}_d, \mathcal{E}_d, \mathbf{A}_d)$, where each edge $\{i, j\} \in \mathcal{E}_p$  denotes a node $u = \{i, j\} \in \mathcal{V}_d$, and two dual nodes $u=\{i, j\}$ and $w=\{k, l\}$ are adjacent if and only if $\{i, j\} \cap \{k, l\} \neq \emptyset$. This construction is invertible and preserves all structural information of the primal graph. Such dual graphs are also referred to as line graphs or adjoint graphs in graph theory \cite{gross2018graph}. 

\textbf{Initialize HR edges.} HR edge features $\mathbf{E}_h \in \mathbb{R}^{n_h \times n_h}$ are initialized as the dot product of incident node features: $\mathbf{E}_{h,ij}=\mathbf{\hat{X}}_{h,i} \cdot \mathbf{\hat{X}}_{h,j}$. More complex functions can be used to obtain multi-dimensional features if required.

\textbf{Primal2Dual Conversion.} Consider a fully-connected primal graph $\mathcal{G}_p = (\mathcal{V}_h, \mathcal{E}_p, \mathbf{1} - \mathbf{I})$, and define its dual as $\mathcal{G}_d = (\mathcal{V}_d =\mathcal{E}_p, \mathcal{E}_d, \mathbf{A}_d, \mathbf{X}_d)$. Each dual node $u =\{i, j\} \in \mathcal{V}_d$ inherits its feature from the corresponding HR edge: $\mathbf{X}_{d,u} = \mathbf{E}_{h,ij}$. 

\textbf{Dual node representation learning.} Dual node features are updated via messgae passing: $\mathbf{\hat{X}}_d = GNN_d(\mathbf{X}_d, \mathbf{A}_d)$. Due to the bijective mapping between dual nodes and primal edges, this is equivalent to learning HR edge representations.

\textbf{Dual2Primal conversion.} The learned dual node features are mapped back to primal edges as $\mathbf{\hat{E}}_{p,ij}= \mathbf{\hat{X}}_{d,u}$.

\textbf{Graph topology inference.} The HR adjacency matrix is inferred as $\mathbf{\hat{A}}_h = \beta(\mathbf{\hat{E}}_p)$, where $\beta(\cdot)$ denotes min-max normalization. A more complex function may be used when edge features are multi-dimensional. 

\textbf{Computational analysis.} For our fully-connected primal graph, the dual graph has $|\mathcal{V}_d| = |\mathcal{E}|_p = n_h(n_h-1)/2$ nodes and $m = 2(n_h-2)$ neighbors per node, resulting in $|\mathcal{E}_d| = m|\mathcal{V}_d|/2=n_h(n_h-1)(n_h-2)/2$ edges, where the factor of 2 corrects double counting. This gives a sparsity of $\dfrac{2 \times |\mathcal{E}_d|}{|\mathcal{V}_d|^2} = 1 - \dfrac{4(n_h-2)}{n_h(n_h-1)}$ for $\mathbf{A}_d$, exceeding $ 90\%$ for $n_h \geq 39$. This high sparsity enables the use of in-built sparse matrix optimization in deep learning libraries, substantially reducing the computational cost of the dual graph formulation.

\section{Experiments}

\subsection{Node vs. Edge representation learning}

We compare node-based and edge-based representation learning to analyze the trade-off between inductive bias and expressivity in edge prediction. Node-based models predict edge values via dot products between node embeddings, which limits them to functions aligned with this structure. In contrast, edge-based models directly encode pairwise interactions and can represent a wider class of edge functions. We evaluate both approaches by varying graph topology and edge-generating functions to identify the conditions where each performs best.

\textbf{Datasets.} We construct synthetic graphs ($n=16$) inspired by interacting particle systems, where each node represents a particle with a 2D position and mass as features. We use three configurations: \textbf{\emph{D1}} (grid graph with random masses), \textbf{\emph{D2}} (random geometric graph with uniform mass), and \textbf{\emph{D3}} (random geometric graph with random masses). Node coordinates and masses are sampled from $\mathcal{U}(0, 1)$, and edges are created between nodes whose Euclidean distance exceeds a threshold $t=0.3$. 

Edge values are generated using function listed in Table \ref{tab1}, chosen to span a range of dependencies, from functions recoverable via dot products on node features to those requiring explicit modeling of pairwise geometry.

\begin{table}[!t]
\caption{Edge functions for node vs. edge representation learning.\label{tab1}}
\renewcommand{\arraystretch}{1.2} 
\resizebox{\columnwidth}{!}{%
\begin{tabular}{|p{0.5cm}|p{4.0cm}|p{4.3cm}|}
\hline
\textbf{Label} & \textbf{Equation Type} & \textbf{Edge Function, $e_{ij}$} \\
\hline
\textbf{\textit{E1}} & Inverse square law & ${G m_i m_j}/((x_i - x_j)^2 + (y_i - y_j)^2)$ \\

\textbf{\textit{E2}} & Asymmetric rational function & $(A m_i + B m_j)/(x_i^2 + y_j^2)$ \\

\textbf{\textit{E3}} & Symmetric quadratic function & $(x_i - x_j)^2 + (y_i - y_j)^2 + (m_i - m_j)^2$ \\

\textbf{\textit{E4}} & Symmetric polynomial function & $x_i y_i m_i + x_j y_j m_j + x_i y_j + x_j y_i$ \\

\textbf{\textit{E5}} & Asymmetric quadratic function & $x_i^2 + y_i^2 + m_j^2$ \\
\hline
\end{tabular}%
}
\end{table}

\textbf{Models.} We construct four simple models - two each for node and edge representation learning. The \textbf{\emph{Node}} model is a single-layer MPNN inspired by GIN \cite{xu2018powerful}, with node updates $\mathbf{\hat{x}}_i = f_{node}(\mathbf{x}_i + \sum_{j\in \mathcal{N}_i} \mathbf{x}_j)$ and edge prediction $\hat{e}_{ij} = \mathbf{\hat{x}}_i \cdot \mathbf{\hat{x}}_j$, where $f_{node}: \mathbb{R}^3 \mapsto \mathbb{R}^{16} \mapsto \mathbb{R}^{16}$ is a two-layer FFN acting as a universal function approximator \cite{hornik1989multilayer}. The \textbf{\emph{Node Large}} model uses a deeper FFN ($\mathbb{R}^3 \mapsto \mathbb{R}^{16} \mapsto \mathbb{R}^{16} \mapsto \mathbb{R}^1$) that directly regresses edge values from scalar node embeddings. For edge representation learning, we initialize edge features as $\mathbf{e}_{ij}' = [\mathbf{x}_i || \mathbf{x}_j]$, where $||$ denotes concatenation. The \textbf{\emph{Edge}} model predicts $\hat{e}_{ij} = f_{edge}(\mathbf{e}_{ij}')$, where $f_{edge} : \mathbb{R}^6 \mapsto \mathbb{R}^{16} \mapsto \mathbb{R}^{16} \mapsto \mathbb{R}^1$. The \textbf{\emph{Dual Edge}} model applies message passing over the dual graph and predicts: $\hat{e}_{ij} = f_{dual\_edge}(\mathbf{e}_{ij}' + \sum_{\{k,l\} \in \mathcal{N}_{\{i, j\}}} \mathbf{e}_{kl}')$, where $f_{dual\_edge}$ shares the same architecture as $f_{edge}$.

\begin{table}[!t]
\caption{Result comparison of node vs. edge representation learning\label{tab2}}
\centering
\renewcommand{\arraystretch}{1.2} 
\resizebox{\columnwidth}{!}{%
\begin{tabular}{|p{0.2cm}|p{2.1cm}|p{2.1cm}|p{2.1cm}|p{2.1cm}|}
\hline
 & \textbf{\emph{Node}} & \textbf{\emph{Node Large}} & \textbf{\emph{Edge}} & \textbf{\emph{Dual Edge}} \\
\hline
\multicolumn{5}{|c|}{Performance on \textbf{\emph{E1}} across graph datasets} \\
\hline
\textbf{\emph{D1}} & \underline{$\mathbf{0.869 \pm 0.032}$} & $\mathbf{1.136 \pm 0.899}$ & $2.371 \pm 2.087$ & $1.565 \pm 1.317$\\
\textbf{\emph{D2}} & $41.176 \pm 25.567$ & $39.525 \pm 28.190$ & \underline{$\mathbf{33.266 \pm 16.387}$} & $\mathbf{38.221 \pm 23.984}$\\
\textbf{\emph{D3}} & $13.499 \pm 9.805$ & $\mathbf{9.012 \pm 5.058}$ & \underline{$\mathbf{8.696 \pm 5.444}$} & $10.873 \pm 5.928$ \\
\hline
\multicolumn{5}{|c|}{Performance on \textbf{\emph{D3}} across edge types} \\
\hline
\textbf{\emph{E1}} & $13.499 \pm 9.805$ & $\mathbf{9.012 \pm 5.058}$ & \underline{$\mathbf{8.696 \pm 5.444}$} & $10.873 \pm 5.928$\\
\textbf{\emph{E2}} & $26.611 \pm 9.176$ & $\mathbf{26.304 \pm 5.800}$ & \underline{$\mathbf{24.702 \pm 6.480}$} & $26.991 \pm 10.280$\\
\textbf{\emph{E3}} & $0.305 \pm 0.014$ & $0.325 \pm 0.037$ & \underline{$\mathbf{0.196 \pm 0.182}$} & $\mathbf{0.249 \pm 0.073}$ \\
\textbf{\emph{E4}} & \underline{$\mathbf{0.485 \pm 0.039}$} & $0.663 \pm 0.391$ & $0.640 \pm 0.710$ & $\mathbf{0.639 \pm 0.275}$ \\
\textbf{\emph{E5}} & \underline{$\mathbf{0.637 \pm 0.036}$} & $0.898 \pm 0.500$ & $0.821 \pm 0.934$ & $\mathbf{0.779 \pm 0.419}$ \\
\hline
\end{tabular}%
}
\end{table}

\textbf{Evaluation and results.} We evaluate model performances in two settings: (1) fixing the edge function to the inverse square law (\textbf{\emph{E1}}) while varying graph structure (\textbf{\emph{D1}}–\textbf{\emph{D3}}), and (2) fixing the graph structure to the most randomized configuration \textbf{\emph{D3}} while varying the edge function (\textbf{\emph{E1}}–\textbf{\emph{E5}}). All models are trained via mean squared loss (MSE), which provides a smooth optimization landscape. Test performance is reported using mean absolute error (MAE) between true and predicted edge values in Table \ref{tab2}.

In the first setting, node-based models outperform edge-based models on \textbf{\emph{D1}}, where inter-node distances are constant and \textbf{\emph{E1}} reduces to the dot product between masses, an inductive bias easily captured by the node-based models. On \textbf{\emph{D2}}, where masses are uniform and edge values depend only on inverse squared distance, node-based models underperform due to their inability to encode this geometric relation via dot products. On \textbf{\emph{D3}}, where both masses and distances vary, performance is comparable, indicating a compensatory effect between numerator and denominator terms in \textbf{\emph{E1}}.

In the second setting, node and edge-based models perform similarly on \textbf{\emph{E1}} and \textbf{\emph{E2}}, benefitting from partial error compensation between numerator and denominator. In contrast, \textbf{\emph{E3}} depends exclusively on squared distance between the node features, which cannot be encoded by node-dot products. Hence, edge-based models outperform in this case. \textbf{\emph{E4}} and \textbf{\emph{E5}} are polynomial in node features and can be approximated through nonlinear projections into higher-dimensional feature spaces, resulting in comparable performance across model types.

Two experimental limitations should be noted. First, stochastic data generation occasionally produces nodes in close proximity, yielding large edge values that skew training under MSE loss. To mitigate this, all experiments are averaged over 15 random seeds. Second, model hyperparameters were not tuned for optimal performance, as the goal is to highlight the trade-off between inductive bias and representational expressivity.

\subsection{Graph super-resolution on simulated datasets}\label{exp:simulated}

To evaluate the proposed frameworks, we construct twelve simulated datasets that reflect diverse graph topologies and controlled LR–HR relationships. Each dataset is generated by first sampling an HR graph from a traditional graph generative model, then deriving its LR counterpart using TopK pooling based on different graph-theoretic measures. This enables a systematic comparison of fourteen ablated models, isolating the relative effectiveness of node super-resolution and edge inference modules under topologically varying scenarios.

\textbf{Simulated Datasets.} We synthetize HR graphs ($n_h=128$) using three traditional graph generation models: (1) Stochastic Block Model (\textbf{\emph{SBM}}) \cite{lee2019review}, which assigns nodes to clusters and probabilistically connects them based on a block-wise connectivity matrix, mimicking modular structures observed in social or biological networks; (2) Barabási-Albert (\textbf{\emph{BA}}) model \cite{barabasi1999emergence}, which grows the graph via preferential attachment, producing scale-free degree distributions commonly found in internet topology and citation networks; and (3) Watts-Strogatz (\textbf{\emph{WS}}) model \cite{watts1998collective}, which rewires a regular ring lattice with a given probability to generate graphs with high clustering and short path lengths, resembling small-world networks such as neural or power grids. Node features are computed using Node2Vec \cite{grover2016node2vec}, which learns embeddings via biased random walks that capture both local and global structure. Edge weights, used as adjacency matrix values, are computed as the Pearson correlation between node feature vectors. 

We create LR graphs using TopK pooling \cite{cangea2018towards} on the HR graphs. Specifically, nodes are ranked according to one of four measures: degree centrality, betweenness centrality, clustering coefficient, or participation coefficient. The top $K=n_l=64$ nodes are retained along with their induced subgraph to form the LR graphs. Together, this results in twelve topologically diverse datasets to benchmark graph super-resolution models. 

\begin{table}[!t]
\caption{Benchmarking results on simulated datasets\label{tab3}}
\centering
\renewcommand{\arraystretch}{1.2} 
\resizebox{\columnwidth}{!}{%
\begin{tabular}{|p{2.2cm}|p{2.0cm}|p{2.0cm}|p{2.0cm}|p{2.0cm}|}
\hline
 \emph{TopK metric} & \textbf{\emph{Degree}} & \textbf{\emph{Betweeness}} & \textbf{\emph{Clustering}} & \textbf{\emph{Participation}} \\
\hline
\multicolumn{5}{|c|}{Performance on \textbf{\emph{SBM}} datasets} \\
\hline
\textbf{\emph{MT}} 
        & $2.841 \pm 0.123$ 
        & $2.712 \pm 0.204$ 
        & $2.591 \pm 0.087$ 
        & $2.784 \pm 0.062$ \\
\hline    
\textbf{\emph{Bi-LC}}
        & \underline{$\mathbf{2.495 \pm 0.109}$} 
        & $\mathbf{2.603 \pm 0.021}$ 
        & $\mathbf{2.463 \pm 0.061}$ 
        & $\mathbf{2.570 \pm 0.061}$\\
        
\textbf{\emph{Bi-LC}}$_{\textit{fixed}}$ 
        & $2.518 \pm 0.119$ 
        & $2.626 \pm 0.015$ 
        & $2.574 \pm 0.091$ 
        & $2.578 \pm 0.076$  \\
        
\textbf{\emph{Bi-LC}}$_{\textit{learned}}$ 
        & $2.595 \pm 0.139$ 
        & $2.678 \pm 0.049$ 
        & $2.574 \pm 0.051$ 
        & $2.592 \pm 0.072$ \\
\hline
\textbf{\emph{Bi-MP}} 
        & $2.548 \pm 0.103$ 
        & $2.685 \pm 0.030$ 
        & $2.494 \pm 0.057$ 
        & $2.592 \pm 0.083$ \\

\textbf{\emph{Bi-MP}}$_{\textit{fixed}}$ 
        & $\mathbf{2.511 \pm 0.117}$ 
        & \underline{$\mathbf{2.594 \pm 0.009}$} 
        & \underline{$\mathbf{2.463 \pm 0.039}$} 
        & $2.572 \pm 0.103$  \\

\textbf{\emph{Bi-MP}}$_{\textit{learned}}$ 
        & $2.523 \pm 0.123$ 
        & $2.659 \pm 0.066$ 
        & $2.553 \pm 0.084$ 
        & $2.691 \pm 0.090$  \\
\hline
\textbf{\emph{Dual MT}} 
        & $3.384 \pm 0.962$ 
        & $3.350 \pm 1.059$ 
        & $2.600 \pm 0.044$ 
        & $3.437 \pm 1.214$ \\
\hline  
\textbf{\emph{Dual Bi-LC}}
        & $2.558 \pm 0.136$ 
        & $2.637 \pm 0.028$ 
        & $2.511 \pm 0.063$ 
        & $2.994 \pm 0.721$\\
        
\textbf{\emph{Dual Bi-LC}}$_{\textit{fixed}}$ 
        & $3.286 \pm 0.565$ 
        & $2.887 \pm 0.307$ 
        & $2.615 \pm 0.166$ 
        & $2.916 \pm 0.132$ \\
        
\textbf{\emph{Dual Bi-LC}}$_{\textit{learned}}$ 
        & $2.589 \pm 0.150$ 
        & $3.404 \pm 0.977$ 
        & $3.066 \pm 0.600$ 
        & $2.750 \pm 0.129$  \\
\hline
\textbf{\emph{Dual Bi-MP}} 
        & $2.562 \pm 0.160$ 
        & $2.660 \pm 0.069$ 
        & $2.524 \pm 0.016$ 
        & \underline{$\mathbf{2.539 \pm 0.044}$}  \\

\textbf{\emph{Dual Bi-MP}}$_{\textit{fixed}}$ 
        & $2.601 \pm 0.119$ 
        & $2.668 \pm 0.032$ 
        & $2.676 \pm 0.143$ 
        & $2.625 \pm 0.019$ \\

\textbf{\emph{Dual Bi-MP}}$_{\textit{learned}}$ 
        & $2.543 \pm 0.093$ 
        & $2.629 \pm 0.017$ 
        & $2.493 \pm 0.088$ 
        & $2.694 \pm 0.220$ \\
\hline
\multicolumn{5}{|c|}{Performance on \textbf{\emph{BA}} datasets} \\
\hline
\textbf{\emph{MT}} 
        & $1.761 \pm 0.107$ 
        & $2.162 \pm 0.364$ 
        & $1.813 \pm 0.075$ 
        & $1.747 \pm 0.055$ \\
\hline    
\textbf{\emph{Bi-LC}}
        & $1.749 \pm 0.033$ 
        & $1.789 \pm 0.038$ 
        & $1.787 \pm 0.054$ 
        & $1.752 \pm 0.028$\\
        
\textbf{\emph{Bi-LC}}$_{\textit{fixed}}$ 
        & $1.738 \pm 0.018$ 
        & $\mathbf{1.767 \pm 0.022}$ 
        & $1.814 \pm 0.054$ 
        & $\mathbf{1.728 \pm 0.037}$  \\
        
\textbf{\emph{Bi-LC}}$_{\textit{learned}}$ 
        & $1.745 \pm 0.018$ 
        & $1.775 \pm 0.058$ 
        & $1.833 \pm 0.028$ 
        & $1.743 \pm 0.044$\\
\hline
\textbf{\emph{Bi-MP}} 
        & $\mathbf{1.721 \pm 0.026}$ 
        & \underline{$\mathbf{1.750 \pm 0.038}$} 
        & $\mathbf{1.753 \pm 0.065}$ 
        & \underline{$\mathbf{1.726 \pm 0.080}$} \\

\textbf{\emph{Bi-MP}}$_{\textit{fixed}}$ 
        & \underline{$\mathbf{1.705 \pm 0.019}$} 
        & $1.795 \pm 0.040$ 
        & \underline{$\mathbf{1.748 \pm 0.093}$} 
        & $1.758 \pm 0.029$ \\

\textbf{\emph{Bi-MP}}$_{\textit{learned}}$ 
        & $1.789 \pm 0.041$ 
        & $1.780 \pm 0.039$ 
        & $1.763 \pm 0.052$ 
        & $1.761 \pm 0.007$   \\
\hline
\textbf{\emph{Dual MT}} 
        & $1.845 \pm 0.047$ 
        & $1.894 \pm 0.050$ 
        & $1.861 \pm 0.096$ 
        & $1.800 \pm 0.031$\\
\hline  
\textbf{\emph{Dual Bi-LC}}
        & $1.775 \pm 0.035$ 
        & $1.802 \pm 0.051$ 
        & $1.775 \pm 0.083$ 
        & $1.732 \pm 0.028$ \\
        
\textbf{\emph{Dual Bi-LC}}$_{\textit{fixed}}$ 
        & $1.984 \pm 0.222$ 
        & $2.446 \pm 0.891$ 
        & $1.966 \pm 0.084$ 
        & $2.659 \pm 1.209$\\
        
\textbf{\emph{Dual Bi-LC}}$_{\textit{learned}}$ 
        & $2.459 \pm 0.880$ 
        & $2.604 \pm 0.712$ 
        & $2.416 \pm 0.905$ 
        & $3.030 \pm 0.924$  \\
\hline
\textbf{\emph{Dual Bi-MP}} 
        & $1.824 \pm 0.018$ 
        & $1.974 \pm 0.091$ 
        & $1.869 \pm 0.024$ 
        & $1.863 \pm 0.130$  \\

\textbf{\emph{Dual Bi-MP}}$_{\textit{fixed}}$ 
        & $1.931 \pm 0.087$ 
        & $1.797 \pm 0.073$ 
        & $1.888 \pm 0.098$ 
        & $1.803 \pm 0.088$ \\

\textbf{\emph{Dual Bi-MP}}$_{\textit{learned}}$ 
        & $1.721 \pm 0.041$ 
        & $1.784 \pm 0.009$ 
        & $1.809 \pm 0.092$ 
        & $1.790 \pm 0.144$\\

\hline
\multicolumn{5}{|c|}{Performance on \textbf{\emph{WS}} datasets} \\
\hline
\textbf{\emph{MT}} 
        & $2.179 \pm 0.132$ 
        & $2.070 \pm 0.061$ 
        & $2.128 \pm 0.053$ 
        & $2.104 \pm 0.100$ \\
\hline    
\textbf{\emph{Bi-LC}}
        & \underline{$\mathbf{1.989 \pm 0.006}$} 
        & $2.007 \pm 0.012$ 
        & $2.002 \pm 0.031$ 
        & $2.022 \pm 0.027$\\
        
\textbf{\emph{Bi-LC}}$_{\textit{fixed}}$ 
        & $2.035 \pm 0.035$ 
        & $2.058 \pm 0.022$ 
        & $2.034 \pm 0.015$ 
        & $2.075 \pm 0.066$  \\
        
\textbf{\emph{Bi-LC}}$_{\textit{learned}}$ 
        & $2.012 \pm 0.020$ 
        & $2.043 \pm 0.043$ 
        & $2.027 \pm 0.067$ 
        & $2.090 \pm 0.046$ \\
\hline
\textbf{\emph{Bi-MP}} 
        & $\mathbf{1.998 \pm 0.020}$ 
        & $\mathbf{2.002 \pm 0.005}$ 
        & $2.016 \pm 0.056$ 
        & $\mathbf{2.013 \pm 0.017}$\\

\textbf{\emph{Bi-MP}}$_{\textit{fixed}}$ 
        & $2.003 \pm 0.014$ 
        & \underline{$\mathbf{1.996 \pm 0.010}$} 
        & $\mathbf{1.998 \pm 0.027}$ 
        & $2.019 \pm 0.028$ \\

\textbf{\emph{Bi-MP}}$_{\textit{learned}}$ 
        & $2.060 \pm 0.083$ 
        & $2.028 \pm 0.035$ 
        & \underline{$\mathbf{1.994 \pm 0.030}$} 
        & \underline{$\mathbf{2.010 \pm 0.025}$} \\
\hline
\textbf{\emph{Dual MT}} 
        & $2.149 \pm 0.011$ 
        & $2.879 \pm 1.197$ 
        & $2.156 \pm 0.027$ 
        & $2.206 \pm 0.085$\\
\hline  
\textbf{\emph{Dual Bi-LC}}
        & $2.027 \pm 0.040$ 
        & $2.007 \pm 0.016$ 
        & $2.009 \pm 0.024$ 
        & $2.422 \pm 0.698$ \\
        
\textbf{\emph{Dual Bi-LC}}$_{\textit{fixed}}$ 
        & $2.615 \pm 0.894$ 
        & $3.025 \pm 1.602$ 
        & $2.466 \pm 0.649$ 
        & $2.320 \pm 0.284$ \\
        
\textbf{\emph{Dual Bi-LC}}$_{\textit{learned}}$ 
        & $2.679 \pm 1.055$ 
        & $2.942 \pm 0.802$ 
        & $2.140 \pm 1.846$ 
        & $2.303 \pm 0.199$ \\
\hline
\textbf{\emph{Dual Bi-MP}} 
        & $2.450 \pm 0.598$ 
        & $2.039 \pm 0.038$ 
        & $2.083 \pm 0.122$ 
        & $2.097 \pm 0.049$\\

\textbf{\emph{Dual Bi-MP}}$_{\textit{fixed}}$ 
        & $2.270 \pm 0.381$ 
        & $2.394 \pm 0.414$ 
        & $2.432 \pm 0.683$ 
        & $2.417 \pm 0.622$ \\

\textbf{\emph{Dual Bi-MP}}$_{\textit{learned}}$ 
        & $2.022 \pm 0.048$ 
        & $2.970 \pm 1.666$ 
        & $2.427 \pm 0.710$ 
        & $2.041 \pm 0.049$ \\
\hline
\end{tabular}%
}
\end{table}

\textbf{Models.} We construct fourteen ablated models by varying the node super-resolution module $\mathcal{S}_{\mathcal{V}}$ and the edge inference module $\mathcal{S}_{\mathcal{E}}$. The first seven use dot-product $\mathcal{S}_{\mathcal{E}}$ and differ in their choice of  $\mathcal{S}_{\mathcal{V}}$ as described in section \ref{section:bi-sr}: \textbf{\emph{MT}} is the baseline model; \textbf{\emph{Bi-MP}} and \textbf{\emph{Bi-LC}} apply bipartite message passing and linear combination without HR node representation learning; \textbf{\emph{Bi-MP}}$_{\textit{fixed}}$, \textbf{\emph{Bi-MP}}$_{\textit{learned}}$, \textbf{\emph{Bi-LC}}$_{\textit{fixed}}$, and \textbf{\emph{Bi-LC}}$_{\textit{learned}}$ extend the latter by including HR node representation learning via either fixed or learnable computational domain. The remaining seven substitute the dot product $\mathcal{S}_{\mathcal{E}}$ with our dual graph formulation, yielding \textbf{\emph{Dual MT}}, \textbf{\emph{Dual Bi-MP}}, \textbf{\emph{Dual Bi-LC}}, \textbf{\emph{Dual Bi-MP}}$_{\textit{fixed}}$, \textbf{\emph{Dual Bi-MP}}$_{\textit{learned}}$, \textbf{\emph{Dual Bi-LC}}$_{\textit{fixed}}$, and \textbf{\emph{Dual Bi-LC}}$_{\textit{learned}}$. All GNNs in the $\mathcal{S}_{\mathcal{V}}$ and $\mathcal{S}_{\mathcal{E}}$ modules use a single-layer graph transformer. Further architectural details are provided in Appendix \ref{appendix:sim}.

\textbf{Evaluation and results.}
All models are trained to minimize the mean absolute error (MAE) between predicted and ground-truth HR edge weights, and evaluated using three-fold cross-validation. Although synthetically generated, these datasets present non-trivial challenges. For example, TopK pooling based on degree centrality may omit entire HR clusters in \textbf{\emph{SBM}} graphs, making it difficult to infer edge weights for the missing clusters. 

Performance results are summarized in Table \ref{tab3}. Models using bipartite message passing consistently outperform others, especially on \textbf{\emph{BA}} and \textbf{\emph{WS}} datasets, where HR graphs resemble extrapolated versions of their LR counterparts. Here, message passing may help propagate structural relationships across scales. On \textbf{\emph{SBM}} datasets, performance gap between bipartite message passing and linear combination is narrower. Here, \textbf{\emph{Bi-LC}}'s flexibility aids in recovering edges from underrepresented clusters. Models with the dual graph formulation do not improve over dot-product variants, because Pearson correlation, used to compute edge weights, is symmetric and can be approximated via dot products of learned node embeddings.

A limitation of our experimental design is its reliance on TopK pooling with traditional graph measures to define LR-HR mappings. While effective for controlled evaluation of specific topological structures, these are insufficient to capture complex and non-hierarchical relationships in real-world graphs.  

\subsection{Brain graph super-resolution}\label{section:brain-sr}

We evaluate our graph super-resolution frameworks on a real-world network neuroscience dataset, aiming to recover high-resolution (HR) connectomes from low-resolution (LR) counterparts. We benchmark sixteen models, including fourteen ablated and two additional baselines, using eight graph-theoretic measures that capture critical aspects of network topology and brain connectivity. Our results highlight the effectiveness of our frameworks, particularly the dual graph formulation, in capturing both local and global brain network structure.

\textbf{Connectomic Dataset.} We use the Southwest University Longitudinal Imaging Multimodal (SLIM) dataset \cite{liu2017longitudinal}, which provides multimodal neuroimaging data for 167 subjects, including resting-state fMRI. For each subject, we generate brain connectivity matrices at two resolutions using different parcellation schemes: the Dosenbach atlas \cite{dosenbach2010prediction} for LR graphs ($n_l = 160$) and the Shen atlas \cite{shen2013groupwise} for HR graphs ($n_h = 268$). These functional connectomes represent inter-regional neural activity correlations and serve as weighted adjacency matrices $\mathbf{A}_l$ and $\mathbf{A}_h$ for LR and HR graphs, respectively. Following prior work \cite{mhiri2021non}, we initialize node features as $\mathbf{X}_l = \mathbf{A}_l$ and $\mathbf{X}_h = \mathbf{A}_h$.

\begin{table*}[!t]
\caption{Performance on brain graph super-resolution\label{tab4}}
\renewcommand{\arraystretch}{1.2}
\resizebox{\textwidth}{!}{%
\begin{tabular}{|p{2.2cm}|p{2.0cm}|p{2.0cm}|p{2.0cm}|p{2.0cm}|p{2.0cm}|p{2.0cm}|p{2.0cm}|p{2.0cm}|}
\hline
& \makecell{\textbf{\emph{Edge Weights}} \\ ($10^1$)} 
& \makecell{\textbf{\emph{Betweenness}} \\ ($10^4$)} 
& \makecell{\textbf{\emph{Closeness}} \\ ($10^1$)} 
& \makecell{\textbf{\emph{Eigenvector}} \\ ($10^3$)} 
& \makecell{\textbf{\emph{Degree}} \\ ($10^0$)} 
& \makecell{\textbf{\emph{Participation}} \\ ($10^1$)} 
& \makecell{\textbf{\emph{Clustering}} \\ ($10^2$)} 
& \makecell{\textbf{\emph{Small Worldness}} \\ ($10^2$)} 
\\
\hline
\textbf{\emph{IMAN}}$_{adapted}$
& $1.725 \pm 0.074$
& $7.695 \pm 0.159$
& $1.590 \pm 0.028$
& $7.507 \pm 0.096$
& $54.778 \pm 1.170$
& $6.850 \pm 0.091$ 
& $14.006 \pm 0.318$ 
& $8.360 \pm 0.243$ \\

\textbf{\emph{Autoencoder}}
& $1.381 \pm 0.062$
& $7.608 \pm 0.204$ 
& $1.520 \pm 0.025$ 
& $7.179 \pm 0.083$
& $51.697 \pm 1.038$ 
& $5.552 \pm 1.450$ 
& $14.193 \pm 0.437$ 
& $8.260 \pm 0.336$ \\

\hline
\textbf{\emph{MT}} 
& \underline{$\mathbf{1.350 \pm 0.066}$} 
& $7.562 \pm 0.152$ 
& $1.513 \pm 0.033$ 
& $7.155 \pm 0.124$ 
& $51.555 \pm 1.458$ 
& $5.255 \pm 0.883$ 
& $14.128 \pm 0.286$ 
& $8.126 \pm 0.289$ \\

\hline
\textbf{\emph{Bi-LC}}
& $1.528 \pm 0.021$ 
& $7.693 \pm 0.159$ 
& $1.590 \pm 0.028$ 
& $7.507 \pm 0.096$ 
& $54.771 \pm 1.170$ 
& $6.858 \pm 0.173$ 
& $14.003 \pm 0.318$ 
& $8.362 \pm 0.240$ \\

\textbf{\emph{Bi-LC}}$_{\textit{fixed}}$
& $1.507 \pm 0.051$ 
& $7.693 \pm 0.159$ 
& $1.590 \pm 0.028$ 
& $7.506 \pm 0.096$ 
& $54.771 \pm 1.170$ 
& $6.836 \pm 0.096$ 
& $14.103 \pm 0.318$ 
& $8.350 \pm 0.244$ \\

\textbf{\emph{Bi-LC}}$_{\textit{learned}}$
& $1.523 \pm 0.055$ 
& $7.693 \pm 0.159$ 
& $1.590 \pm 0.028$ 
& $7.506 \pm 0.096$ 
& $54.771 \pm 1.170$ 
& $6.822 \pm 0.106$ 
& $14.103 \pm 0.318$ 
& $8.358 \pm 0.243$ \\

\hline
\textbf{\emph{Bi-MP}}
& $1.455 \pm 0.031$ 
& $7.658 \pm 0.208$ 
& $1.578 \pm 0.043$ 
& $7.453 \pm 0.182$ 
& $54.341 \pm 1.730$ 
& $6.410 \pm 0.849$ 
& $13.956 \pm 0.369$ 
& $8.331 \pm 0.287$ \\

\textbf{\emph{Bi-MP}}$_{\textit{fixed}}$
& $1.428 \pm 0.052$ 
& $7.588 \pm 0.156$ 
& $1.551 \pm 0.039$ 
& $7.325 \pm 0.127$ 
& $53.324 \pm 1.650$ 
& $5.090 \pm 0.837$ 
& $13.916 \pm 0.272$ 
& $8.254 \pm 0.243$ \\

\textbf{\emph{Bi-MP}}$_{\textit{learned}}$
& $1.443 \pm 0.048$ 
& $7.586 \pm 0.192$ 
& $1.554 \pm 0.040$ 
& $7.342 \pm 0.169$ 
& $53.521 \pm 1.651$ 
& $5.576 \pm 0.766$ 
& $13.866 \pm 0.353$ 
& $8.270 \pm 0.268$ \\

\hline 
\textbf{\emph{Dual MT}} 
& $1.458 \pm 0.153$ 
& $5.888 \pm 1.914$ 
& $1.133 \pm 0.442$ 
& $7.360 \pm 0.957$ 
& $38.991 \pm 13.900$ 
& $3.401 \pm 3.172$ 
& $11.953 \pm 5.235$ 
& $5.873 \pm 3.221$ \\

\hline
\textbf{\emph{Dual Bi-LC}}
& $1.515 \pm 0.293$ 
& $5.567 \pm 2.235$ 
& \underline{$\mathbf{0.812 \pm 0.123}$} 
& $6.736 \pm 1.172$ 
& \underline{$\mathbf{31.948 \pm 5.635}$} 
& \underline{$\mathbf{1.330 \pm 0.159}$} 
& \underline{$\mathbf{7.779 \pm 2.068}$} 
& $\mathbf{3.886 \pm 1.847}$ \\

\textbf{\emph{Dual Bi-LC}}$_{\textit{fixed}}$
& $1.609 \pm 0.176$ 
& \underline{$\mathbf{5.376 \pm 0.071}$} 
& $1.030 \pm 0.012$ 
& $6.560 \pm 0.172$ 
& $37.555 \pm 0.806$ 
& $\mathbf{1.382 \pm 0.080}$ 
& $\mathbf{9.718 \pm 0.358}$ 
& $4.086 \pm 1.036$ \\

\textbf{\emph{Dual Bi-LC}}$_{\textit{learned}}$
& $1.646 \pm 0.086$ 
& $7.318 \pm 0.713$ 
& $1.249 \pm 0.366$ 
& $7.504 \pm 0.556$ 
& $45.300 \pm 10.049$ 
& $3.615 \pm 2.714$ 
& $11.874 \pm 2.623$ 
& $7.188 \pm 1.320$ \\

\hline
\textbf{\emph{Dual Bi-MP}}
& $1.488 \pm 0.143$ 
& $\mathbf{5.446 \pm 0.927}$ 
& $\mathbf{0.939 \pm 0.059}$ 
& $6.469 \pm 0.370$ 
& $\mathbf{34.298 \pm 2.567}$ 
& $1.461 \pm 0.204$ 
& $10.064 \pm 1.623$ 
& \underline{$\mathbf{3.451 \pm 0.696}$} \\

\textbf{\emph{Dual Bi-MP}}$_{\textit{fixed}}$
& $1.554 \pm 0.185$ 
& $5.747 \pm 0.848$ 
& $1.031 \pm 0.147$ 
& \underline{$\mathbf{6.373 \pm 0.411}$} 
& $53.324 \pm 1.650$ 
& $5.090 \pm 0.837$ 
& $13.916 \pm 0.272$ 
& $8.254 \pm 0.243$ \\

\textbf{\emph{Dual Bi-MP}}$_{\textit{learned}}$
& $\mathbf{1.373 \pm 0.039}$ 
& $5.742 \pm 0.913$ 
& $1.046 \pm 0.128$ 
& $\mathbf{6.379 \pm 0.276}$ 
& $37.527 \pm 3.782$ 
& $1.440 \pm 0.233$ 
& $10.714 \pm 2.245$ 
&$5.322 \pm 1.068$ \\

\hline
\end{tabular}%
}
\end{table*}

\textbf{Models.} We evaluate the fourteen ablated models from Section \ref{exp:simulated}, along with two additional baselines. The first is \textbf{\emph{IMAN}}$_{adapted}$, a modified version of IMANGraphNet \cite{mhiri2021non}, the current state-of-the-art in graph super-resolution. To address memory issues caused by NNConv layers \cite{simonovsky2017dynamic}, we introduce a linear projection that reduces the dimensionality of input features $\mathbf{X}_l$ prior to NNConv and restore the original dimension afterward. The second is a GNN-based \textbf{\emph{Autoencoder}}, inspired by iterative up- and down-sampling in image super-resolution\cite{haris2018deep}, that predicts the HR graph from the LR input and reconstructs the LR graph from the predicted HR. Both encoder and decoder use a single-layer graph transformer, and the model is trained using the sum of reconstruction losses on both resolutions.

\textbf{Evaluation and results.}
Performance of all sixteen models is measured using the mean absolute error (MAE) between the predicted and ground-truth edge weights in $\mathbf{A}_h$, as well as seven graph-theoretic measures that characterize topological properties of brain networks: Betweenness Centrality (\textbf{\emph{Betweenness}}), Closenness Centrality (\textbf{\emph{Closenness}}), Eigenvector Centrality (\textbf{\emph{Eigenvector}}), Node Degree Centrality (\textbf{\emph{Degree}}), Participation Centrality (\textbf{\emph{Participation}}), Clustering Coefficient (\textbf{\emph{Clustering}}), and Small Worldness (\textbf{\emph{Small Worldness}}). These measures are selected for their relevance in neuroscience, offering insights into network integration, segregation, and robustness.

\textbf{\emph{Degree}} quantifies the number of incident connections to a given node and serves as an indicator of network resilience \cite{achard2006resilient}. \textbf{\emph{Betweenness}} measures the proportion of shortest paths passing through a node, identifying potential hubs or bridge nodes between disparate brain regions \cite{rubinov2010complex}. \textbf{\emph{Closenness}} captures the average shortest path from a node to all others, reflecting the efficiency of information transfer. \textbf{\emph{Eigenvector}} assigns importance to a node based on its connections to other high-scoring nodes, indicating hierarchical influence \cite{lorenzini2023eigenvector}. \textbf{\emph{Participation}} measures the diversity of a node’s intermodular connections, while \textbf{\emph{Clustering}} quantifies the density of local cliques, both of which are critical for understanding network segregation and information processing within specialized brain subsystems \cite{gamboa2014working}. Finally, \textbf{\emph{Small Worldness}} is defined as the ratio of normalized characteristic path length to mean clustering coefficient, capturing the balance between local specialization and global integration during information processing \cite{watts1998collective}.

Table \ref{tab4} reports the MAE across eight evaluation measures. Models incorporating the dual graph formulation achieve the best performance across all topological measures, substantially outperforming  \textbf{\emph{IMAN}}$_{adapted}$ and \textbf{\emph{Autoencoder}} baselines. Among bipartite architectures, message-passing consistently outperforms linear combination when used in isolation, but the gap narrows in dual graph variants. This suggests that the dual formulation provides a robust mechanism to refine initially learned edge features, uplifting the performance of linear combination models. Notably, \textbf{\emph{MT}} remains competitive, indicating that the \textbf{{Bi-SR}} framework alone does not guarantee improvement on this particular dataset. 

\section{Conclusion}

We introduced Bi-SR and DEFEND, two GNN-agnostic frameworks designed to overcome fundamental structural limitations in existing graph super-resolution methods. Bi-SR constructs a bipartite graph between low-resolution (LR) and high-resolution (HR) nodes to enable structurally-aware node super-resolution that preserves permutation invariance. DEFEND creates an invertible mapping between HR edges and dual nodes, enabling efficient and expressive edge inference via standard node-based GNNs. A summary of their architectural advantages over prior methods is presented in Table \ref{tab0}.

For rigorous evaluation, we introduced twelve simulated datasets spanning diverse topologies and LR–HR mappings. Our models consistently outperformed existing methods across these benchmarks. On real brain connectome data, they achieved state-of-the-art results on seven topological measures critical for neuroscientific analysis.

Note that this work focuses on principled design rather than architecture-specific optimization. Despite using simple GNN architectures, our frameworks demonstrate strong performance, highlighting the importance of structural inductive bias. Future work includes extending these methods to larger and more complex graphs, and developing standardized benchmarks for real-world graph super-resolution.

\appendices

\section{GNN architecture}

All GNNs used in this work employ graph transformer layer \cite{shi2020masked}, which extends multi-head self-attention to graph data. We adopt this layer due to its strong expressivity and better computational efficiency compared to standard alternatives like GCNConv \cite{kipf2016semi} and NNConv\cite{simonovsky2017dynamic}, both widely used in graph super-resolution. Each layer is followed by GraphNorm \cite{cai2021graphnorm} to stabilize training, and ReLU activation. 

\section{Complexity Analysis}

In \textbf{Bi-SR}, message passing over the bipartite graph with $n_l$ LR and $n_h$ HR nodes involves $\mathcal{O}(n_l n_h)$ edges, leading to a time complexity of $\mathcal{O}(n_l n_h d)$ for feature dimension $d$. Optional LR and HR node refinement adds $\mathcal{O}(n_l^2 d)$ and $\mathcal{O}(n_h^2 d)$. Memory usage includes bipartite edge storage and node features, totaling $\mathcal{O}(n_l n_h + (n_l + n_h)d)$. Overall, Bi-SR scales quadratically with HR nodes.

In \textbf{DEFEND}, a fully connected HR graph contains $\mathcal{O}(n_h^2)$ edges, each becoming a node in the dual graph. Dual edges connect if their corresponding HR edges share a node, yielding $\mathcal{O}(n_h^3)$ dual edges. Message passing thus incurs a time complexity of $\mathcal{O}(n_h^3 d_e)$, where $d_e$ is the edge feature dimension. Memory usage is $\mathcal{O}(n_h^2 d_e + n_h^3)$ for storing dual node features and adjacency. As a result, DEFEND scales cubically with HR nodes.

\begin{table}[!t]
\centering
\caption{Time and Memory Complexity of Bi-SR and DEFEND
\label{tabc}}
\begin{tabular}{|c|c|c|}
\hline
\textbf{Method} & \textbf{Time Complexity} & \textbf{Memory Complexity} \\
\hline
\textbf{Bi-SR} & $\mathcal{O}(n_l n_h d + n_l^2 d + n_h^2 d)$ & $\mathcal{O}(n_l n_h + (n_l + n_h)d)$ \\
\hline
\textbf{DEFEND} & $\mathcal{O}(n_h^3 d_e)$ & $\mathcal{O}(n_h^2 d_e + n_h^3)$ \\
\hline
\end{tabular}
\end{table}

\section{Sensitivity Analysis of HR node initialization}

We evaluate the sensitivity of our bipartite message passing (\textbf{\emph{Bi-MP}}) framework to the random initialization strategy introduced in Section \ref{section:bi-sr}, where HR node features are initialized from a uniform distribution $\mathcal{U}(0, a)$ with $a$ as a positive scalar. To assess robustness, we re-run brain graph super-resolution experiments 15 times for six \textbf{\emph{Bi-MP}}-based models: \textbf{\emph{Bi-MP}}, \textbf{\emph{Bi-MP}}$_{\textit{fixed}}$, \textbf{\emph{Bi-MP}}$_{\textit{learned}}$, \textbf{\emph{Dual Bi-MP}}, \textbf{\emph{Dual Bi-MP}}$_{\textit{fixed}}$, \textbf{\emph{Dual Bi-MP}}$_{\textit{learned}}$. Each model is tested across three initialization scales: $\mathcal{U}(0, 1)$, $\mathcal{U}(0, 10)$, and $\mathcal{U}(0, 100)$, using five random seeds per scale.

To quantify robustness, we define a relative sensitivity metric:

\begin{equation}
    s_{rel} = \dfrac{max(\{\sigma_{sm} | s\in scales, m \in models_{Bi-MP}\})}{\sigma_{all\_models}}
\end{equation}

where $\sigma_{sm}$ is the standard deviation of the mean MAE (averaged over seeds) for model $m$ at scale $s$, and $\sigma_{all\_models}$ is the standard deviation of MAE scores across all sixteen models from Section \ref{section:brain-sr}. This metric captures the worst-case sensitivity of Bi-MP models relative to the overall variability observed across models.

\begin{table*}[!t]
\caption{Sensitivity of HR node initialization in Bipartite Message Passing\label{tab5}}
\renewcommand{\arraystretch}{1.2}
\resizebox{\textwidth}{!}{%
\begin{tabular}{|p{2.1cm}|p{0.3cm}|p{1.9cm}|p{1.9cm}|p{1.9cm}|p{1.9cm}|p{1.9cm}|p{1.9cm}|p{1.9cm}|p{2.0cm}|}
\hline
& $s$
& \makecell{\textbf{\emph{Edge Weights}} \\ ($10^1$)} 
& \makecell{\textbf{\emph{Betweenness}} \\ ($10^4$)} 
& \makecell{\textbf{\emph{Closeness}} \\ ($10^1$)} 
& \makecell{\textbf{\emph{Eigenvector}} \\ ($10^3$)} 
& \makecell{\textbf{\emph{Degree}} \\ ($10^0$)} 
& \makecell{\textbf{\emph{Participation}} \\ ($10^1$)} 
& \makecell{\textbf{\emph{Clustering}} \\ ($10^2$)} 
& \makecell{\textbf{\emph{Small Worldness}} \\ ($10^2$)} 
\\
\hline
\multicolumn{10}{|c|}{Across models without dual graph formulation} \\
\hline
\multirow{3}{*}{\textbf{\emph{Bi-MP}}}
& 1   & $1.488 \pm 0.025$ & $7.693 \pm 0.000$ & $1.590 \pm 0.000$ & $7.506 \pm 0.000$ & $54.771 \pm 0.000$ & $6.838 \pm 0.023$ & $14.003 \pm 0.000$ & $8.360 \pm 0.000$ \\
& 10  & $1.537 \pm 0.012$ & $7.611 \pm 0.013$ & $1.558 \pm 0.007$ & $7.357 \pm 0.032$ & $53.557 \pm 0.299$ & $5.920 \pm 0.423$ & $13.938 \pm 0.003$ & $8.282 \pm 0.018$ \\
& 100 & $1.614 \pm 0.026$ & $7.665 \pm 0.022$ & $1.580 \pm 0.008$ & $7.460 \pm 0.038$ & $54.414 \pm 0.304$ & $6.692 \pm 0.185$ & $13.974 \pm 0.002$ & $8.333 \pm 0.020$ \\
\hline

\multirow{3}{*}{\textbf{\emph{Bi-MP}}$_{\textit{fixed}}$}
& 1   & $1.417 \pm 0.018$ & $7.596 \pm 0.028$ & $1.555 \pm 0.013$ & $7.350 \pm 0.061$ & $53.511 \pm 0.523$ & $5.489 \pm 0.642$ & $13.907 \pm 0.035$ & $8.270 \pm 0.032$ \\
& 10  & $1.448 \pm 0.029$ & $7.589 \pm 0.034$ & $1.548 \pm 0.014$ & $7.314 \pm 0.064$ & $53.170 \pm 0.569$ & $5.155 \pm 0.717$ & $13.941 \pm 0.020$ & $8.253 \pm 0.041$ \\
& 100 & $1.564 \pm 0.021$ & $7.604 \pm 0.027$ & $1.558 \pm 0.008$ & $7.362 \pm 0.037$ & $53.562 \pm 0.353$ & $5.759 \pm 0.698$ & $13.934 \pm 0.053$ & $8.267 \pm 0.042$ \\
\hline
\multirow{3}{*}{\textbf{\emph{Bi-MP}}$_{\textit{learned}}$}
& 1   & $1.420 \pm 0.024$ & $7.598 \pm 0.045$ & $1.555 \pm 0.021$ & $7.346 \pm 0.095$ & $53.467 \pm 0.829$ & $5.435 \pm 0.881$ & $13.911 \pm 0.030$ & $8.278 \pm 0.037$ \\
& 10  & $1.459 \pm 0.013$ & $7.599 \pm 0.007$ & $1.551 \pm 0.010$ & $7.327 \pm 0.044$ & $53.279 \pm 0.456$ & $5.613 \pm 0.356$ & $13.954 \pm 0.059$ & $8.264 \pm 0.012$ \\
& 100 & $1.566 \pm 0.014$ & $7.617 \pm 0.037$ & $1.563 \pm 0.011$ & $7.383 \pm 0.052$ & $53.780 \pm 0.430$ & $6.166 \pm 0.804$ & $13.924 \pm 0.071$ & $8.289 \pm 0.037$ \\
\hline
$s_{rel}$ & & \textbf{0.060} & \textbf{0.018} & \textbf{0.044} & \textbf{0.039} & \textbf{0.101} & \textbf{0.449} & \textbf{0.015} & \textbf{0.015} \\
\hline
\multicolumn{10}{|c|}{Across models with dual graph formulation} \\
\hline
\multirow{3}{*}{\textbf{\emph{Dual Bi-MP}}}
& 1   & $1.517 \pm 0.054$ & $6.099 \pm 0.474$ & $1.117 \pm 0.105$ & $6.589 \pm 0.256$ & $40.017 \pm 3.009$ & $1.685 \pm 0.143$ & $11.343 \pm 1.067$ & $4.915 \pm 1.144$ \\
& 10  & $1.569 \pm 0.036$ & $5.909 \pm 0.261$ & $1.028 \pm 0.054$ & $6.608 \pm 0.114$ & $38.000 \pm 1.510$ & $1.613 \pm 1.067$ & $10.436 \pm 0.694$ & $4.200 \pm 0.696$ \\
& 100 & $1.585 \pm 0.029$ & $6.018 \pm 0.257$ & $1.152 \pm 0.123$ & $6.668 \pm 0.099$ & $41.664 \pm 3.440$ & $1.646 \pm 0.162$ & $10.978 \pm 0.645$ & $5.608 \pm 1.370$ \\
\hline
\multirow{3}{*}{\textbf{\emph{Dual Bi-MP}}$_{\textit{fixed}}$}
& 1   & $1.569 \pm 0.036$ & $5.965 \pm 0.401$ & $0.938 \pm 0.061$ & $6.580 \pm 0.088$ & $35.079 \pm 1.874$ & $1.533 \pm 0.123$ & $9.844 \pm 0.603$ & $4.265 \pm 1.009$ \\
& 10  & $1.607 \pm 0.021$ & $6.298 \pm 0.125$ & $0.982 \pm 0.036$ & $6.938 \pm 0.240$ & $36.868 \pm 0.871$ & $1.623 \pm 0.134$ & $10.203 \pm 0.296$ & $4.500 \pm 0.402$ \\
& 100 & $1.651 \pm 0.053$ & $5.540 \pm 0.277$ & $1.081 \pm 0.056$ & $6.735 \pm 0.108$ & $39.831 \pm 1.498$ & $1.647 \pm 0.224$ & $11.216 \pm 0.495$ & $3.909 \pm 1.022$ \\
\hline
\multirow{3}{*}{\textbf{\emph{Dual Bi-MP}}$_{\textit{learned}}$}
& 1   & $1.585 \pm 0.029$ & $5.964 \pm 0.273$ & $1.052 \pm 0.043$ & $6.589 \pm 0.305$ & $38.011 \pm 1.209$ & $1.549 \pm 0.120$ & $10.945 \pm 0.360$ & $4.914 \pm 0.607$ \\
& 10  & $1.438 \pm 0.045$ & $6.418 \pm 0.275$ & $1.036 \pm 0.051$ & $6.468 \pm 0.117$ & $37.138 \pm 1.382$ & $1.471 \pm 0.166$ & $10.458 \pm 0.846$ & $5.129 \pm 0.704$ \\
& 100 & $1.567 \pm 0.064$ & $6.692 \pm 0.420$ & $1.201 \pm 0.059$ & $6.835 \pm 0.076$ & $43.085 \pm 1.939$ & $1.709 \pm 0.063$ & $11.952 \pm 0.591$ & $5.275 \pm 0.569$ \\
\hline
$s_{rel}$ & & \textbf{0.133} & \textbf{0.187} & \textbf{0.260} & \textbf{0.126} & \textbf{0.419} & \textbf{0.114} & \textbf{0.231} & \textbf{0.503} \\
\hline
\end{tabular}%
}
\end{table*}

Table \ref{tab5} summarizes the results. All models exhibit low sensitivity to initialization scale. Notably, models without the dual graph formulation show greater robustness, consistent with the observation that higher-capacity models (e.g., dual graph variants) tend to be more sensitive to initialization, especially under limited data.

\section{Experimental Setup and Hyperparameters}

\subsection{Node vs. Edge representation learning}

For \textbf{\emph{E1}}, we use $G = 100.0$ (\textbf{\emph{D1}}), $1.0$ (\textbf{\emph{D2}}, \textbf{\emph{D3}}); for \textbf{\emph{E2}}, we set $A = 10$, $B = -7$. Each experiment is repeated with 15 random seeds. Models are trained using MSE loss, early stopping based on validation loss, and a minimum warmup period. Table~\ref{tab6} gives shared hyperparameters.

\begin{table}[!t]
\centering
\caption{Hyperparameters for node vs. edge experiments
\label{tab6}}
\begin{tabular}{|l|l|r|}
\hline
\textbf{Category} & \textbf{Hyperparameter} & \textbf{Value} \\
\hline
\multirow{2}{*}{Data} 
& Number of nodes & 16 \\
& Samples (train / val / test) & 128 / 32 / 32 \\
\hline
\multirow{3}{*}{Training} 
& Batch size & 16 \\
& Learning rate & 0.001 \\
& Max epochs / Warmup / Patience  & 300 / 10 / 15 \\
\hline
\end{tabular}
\end{table}

\subsection{Graph super-resolution on simulated datasets}\label{appendix:sim}

Each scenario is run using 3-fold cross-validation. For each fold, we split into train/val/test sets and apply early stopping based on validation MAE. Final performance is averaged across folds. Models are trained to minimize MAE between predicted and ground truth HR edge weights. Common hyperparameters are listed in Table~\ref{tab7}.

\begin{table}[!t]
\caption{Hyperparameters for simulated graph super-resolution\label{tab7}}
\renewcommand{\arraystretch}{1.2} 
\resizebox{\columnwidth}{!}{%
\begin{tabular}{|l|l|r|}
\hline
\textbf{Category} & \textbf{Hyperparameter} & \textbf{Value} \\
\hline
\multirow{2}{*}{Data} 
& Samples per scenario & 128 \\
& HR / LR nodes ($n_h$ / $n_l = K$) & 64 / 32 \\
\hline
\multirow{3}{*}{Training} 
& Batch size & 16 \\
& Learning rate & 0.001 \\
& Max epochs / Warmup / Patience & 150 / 15 / 5 \\
\hline
\multirow{3}{*}{Graph Transformer} 
& Hidden dimension & 16 \\
& Attention heads & 4 \\
& Dropout rate & 0.2 \\
\hline
\multirow{3}{*}{\textbf{\emph{SBM}}} 
& Min / Max number of clusters & 2 / 5 \\
& Min / Max intra-cluster connection probability & 0.50 / 0.60 \\
& Min / Max inter-cluster connection probability & 0.01 / 0.10 \\
\hline
\textbf{\emph{BA}} & Min / Max edges per new node & 4 / 8 \\
\hline
\multirow{2}{*}{\textbf{\emph{WS}}} 
& Min / Max nearest neighbors & 4 / 8 \\
& Min / Max rewiring probability  & 0.2 / 0.5 \\
\hline
\multirow{3}{*}{Node2Vec} 
& Node embedding dimension & 8 \\
& HR / LR random walk length & 51 / 26 \\
& Number of random walks per node & 100 \\
\hline
\end{tabular}%
}
\end{table}

\subsection{Brain graph super-resolution}

We follow the same experimental setup as in Section \ref{appendix:sim}, with two modifications. First, we perform a categorical search over learning rates \{0.01, 0.005, 0.001\} and select the best-performing value for each model. Second, we adjust a few hyperparameters: the number of warmup epochs is increased to 30, early stopping patience is reduced to 7, and the hidden dimension in Graph Transformer is set to 32. All others remain unchanged.

\bibliographystyle{IEEEtran}
\bibliography{citations}

\end{document}